\UseRawInputEncoding
\documentclass[9pt,shortpaper,twoside,web]{ieeecolor}
\usepackage{generic}
\usepackage{cite}
\usepackage{amsmath,amssymb,amsfonts}
\usepackage{algorithmic}
\usepackage{graphicx}
\usepackage{textcomp}
\usepackage{algorithmic}
\usepackage{lipsum}
\usepackage[flushleft]{threeparttable}
\usepackage{booktabs,caption}
\usepackage{dblfloatfix}
\usepackage{subcaption}
\usepackage{url}

\def\BibTeX{{\rm B\kern-.05em{\sc i\kern-.025em b}\kern-.08em
    T\kern-.1667em\lower.7ex\hbox{E}\kern-.125emX}}
\begin{document}
\title{Patient-independent Epileptic Seizure Prediction using Deep Learning Models.}

\author{Theekshana~Dissanayake,~Tharindu~Fernando,~\IEEEmembership{Member,~IEEE,}~Simon~Denman,~\IEEEmembership{Member,~IEEE},\\ ~Sridha~Sridharan~\IEEEmembership{Life Senior Member,~IEEE,}~and~Clinton~Fookes,~\IEEEmembership{Senior Member,~IEEE}
\thanks{
Theekshana Dissanayake, Tharindu Fernando, Simon Denman, Sridha Sridharan and Clinton Fookes are from the SAIVT Research group at the Queensland University of Technology, Australia}}
\maketitle

\begin{abstract}
Objective: Epilepsy is one of the most prevalent neurological diseases among humans and can lead to severe brain injuries, strokes, and brain tumors. Early detection of seizures can help to mitigate injuries, and can be used to aid the treatment of patients with epilepsy. The purpose of a seizure prediction system is to successfully identify the pre-ictal brain stage, which occurs before a seizure event. Patient-independent seizure prediction models are designed to offer accurate performance across multiple subjects within a dataset, and have been identified as a real-world solution to the seizure prediction problem. However, little attention has been given for designing such models to adapt to the high inter-subject variability in EEG data. 

Methods: We propose two patient-independent  deep learning architectures with different learning strategies that can learn a global function utilizing data from multiple subjects.

Results: Proposed models achieve state-of-the-art performance for seizure prediction on the CHB-MIT-EEG dataset, demonstrating 88.81\% and 91.54\% accuracy respectively. 

Conclusions: The Siamese model trained on the proposed learning strategy is able to learn patterns related to patient variations in data while predicting seizures. 

Significance: Our models show superior performance for patient-independent seizure prediction, and the same architecture can be used as a patient-specific classifier after model adaptation. We are the first study that employs model interpretation to understand classifier behavior for the task for seizure prediction, and we also show that the MFCC feature map utilized by our models contains predictive biomarkers related to interictal and pre-ictal brain states. 
\end{abstract}


\section{Introduction}\label{sec:introduction}

According to the World Health Organization (WHO), epilepsy is a chronic non-communicable disease of the brain that affects humans of all ages. Around 50 million people in the world suffer from epilepsy, and almost 80\% of those people live in third-world countries \cite{epilepsy--web}. Since epileptic seizures are unpredictable events, they affect the daily lives of sufferers by leading to unexpected accidents and increased mental stress. Early detection of seizures can help to reduce the physical and mental damage caused, and supports early medical diagnosis of seizures \cite{7307237,AHMEDTARISTIZABAL201965,Yuan2019ADetection}.

Electroencephalography (EEG) is commonly used to study variations in brain activity and helps to identify normal and abnormal events occurring in the human brain. In addition, EEG is relatively low-cost which makes it ideal for patients with epilepsy. To accurately determine seizure events, longer duration EEG signals need to be collected, which requires expert monitoring and constrained experimental settings.  

\begin{figure}[h!]
    \centering
    \includegraphics[width=1.0\linewidth]{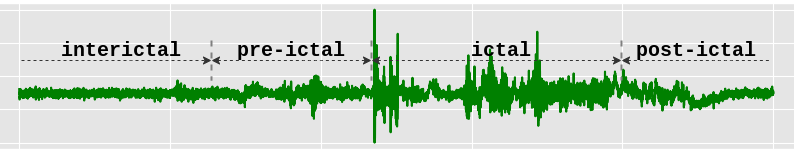}
    \caption{Brain states variation in a typical EEG signal channel. }
    \label{fig:eeg}
\end{figure}

Figure \ref{fig:eeg} illustrates the four typical brain states of an epileptic seizure event. The interictal state refers to the normal brain state of a patient. The brain state before a seizure event is termed the pre-ictal state. This state may last from minutes to hours depending on the subject. The ictal state is the state in which the seizure occurs, and after the seizure event the brain shifts to the post-ictal state. 

Generally, epileptic seizure-related machine learning applications can be categorized into three common types. The first type is epileptic seizure prediction where a classifier is designed to predict a seizure event by identifying the pre-ictal brain state of a given subject \cite{Khan2018FocalNetworks,Daoud2019EfficientLearning}. The second category is seizure detection (or abnormal EEG detection), where a model is designed to classify between seizure (ictal) and non-seizure (interictal) EEG segments \cite{seizuredetection,tbme1}. The third category deals with seizure classification where an EEG seizure sample is classified into the specific seizure type (e.g. focal or non-focal seizures) \cite{seizclsffernando}. Here, in this study, we focus on epileptic seizure prediction considering its importance as an early diagnosis technique.

Seizure prediction deals with designing models to distinguish between pre-ictal and interictal states of the given subject's brain. Since the pre-ictal state duration is subjective \cite{Khan2018FocalNetworks}, pre-ictal duration becomes a design  choice in proposed algorithms. Hence, if the classifier predicts the given EEG signal is pre-ictal, the model is indicating that a seizure will occur within the defined pre-ictal time (duration). The early prediction capability of the designed classifier varies depending on the duration taken as the pre-ictal time. For instance, if the pre-ictal duration is defined as one hour, then the designed classifier has the ability to recognize seizures with a one-hour prediction window \cite{Daoud2019EfficientLearning}. In addition, researchers often consider the interictal brain state to be four hours before or after seizure onset \cite{Daoud2019EfficientLearning}. 

Epileptic seizure prediction can be solved using two approaches:
\begin{enumerate}
    \item \textbf{Patient-independent} studies aim to design a classifier that can recognize seizures across multiple subjects. When designing such models, the entire dataset is utilized, and the objective is to learn a global predictive function that has the ability to perform prediction across multiple subjects in the dataset  \cite{Roy2019DeepReview}. 

    \item \textbf{Patient-specific} studies deal with designing one classifier per subject considering the high inter subject variability in EEG data \cite{Khan2018Cost-sensitiveData}. Here, a single classifier architecture is designed and is fine-tuned for each subject. The final performance of the model is denoted by the average accuracy after training/testing the classifier across all or a selected set of subjects in the dataset \cite{Daoud2019EfficientLearning}. This approach simplifies the problem by focusing on each subject separately when designing a model, but these methods suffer from limited data availability.

\end{enumerate}

It should be noted that designing patient-independent models is a complex task given that EEG data contain high inter subject variability. Acknowledging this, researchers have simplified the problem by designing patient-specific models, however we argue that they do not offer a reliable solution for the problem when a subject in the dataset has fewer recordings. Studying recent literature, some researchers have generated artificial data \cite{artifical} and others have ignored subjects with fewer samples when designing patient-specific models \cite{Daoud2019EfficientLearning}. Therefore, there is an essential need to investigate deep learning techniques  that can learn from data from multiple subjects to effectively address the patient-independent seizure prediction task \cite{LASHGARI2020108885,Roy2019DeepReview}.

The main contribution of this study is the proposal of a patient-independent model for epileptic seizure prediction. Such a method has the capability to learn from subjects with fewer samples while learning from subjects with comparatively higher number of samples uniformly, providing attention to all subjects in the dataset. It should be acknowledged that if the number of recordings available for each patient is consistent across the dataset, and those recordings have longer durations, then designing a patient specific model is a feasible approach. However, often medical datasets comprise a varied number of recordings per subject, and therefore, being able to learn a global function considering the entire dataset is important for real world scenarios \cite{Roy2019DeepReview}. 

Since the medical definition of an epileptic seizure event is uniform for humans, another advantage of learning a global function is it can be used to understand overall patterns in the dataset. Given that existing epileptic seizure studies lack model interpretation \cite{Rasheed2020MachineReview}, the designed model can be used along with a model interpretation algorithm to understand hidden seizure-related patterns in data. Furthermore, designing patient-independent models can be used as a prior step for designing patient-specific seizure classifiers; enabling the creation of models from less data \cite{dataaugment}.  Our main contributions  are  the following: 

\begin{enumerate}

\item We propose two different patient-independent seizure prediction models with a one-hour early prediction window, which out perform the state-of-the-art approaches: \textbf{Model~I}: 88.81($\pm$0.27)\%, \textbf{Model~II}: 91.54($\pm0.17$)\% on the
 the CHB-MIT EEG dataset \cite{Shoeb2009ApplicationTECHNOLOGY}. 

\item We employ model interpretation to understand the input attribution. To the best of our knowledge, this is the first study that uses model interpretation to analyze the channel-level input contribution of models designed for epileptic seizure prediction. 

\item We examine the change in predictive characteristic (bio-marker) when the brain state shifts from the interictal stage to the pre-ictal stage using a probabilistic approach.  

\item We demonstrate how our Siamese network can be transferred to a patient specific model with almost 97\% average accuracy. 
\end{enumerate}

The rest of the paper is organized as follows. In Section \ref{sec:relatedw}, we discuss recent investigations related to epileptic seizure prediction. In Section \ref{sec: methlgy}, we explain the dataset used for the study and deep learning strategies adopted. In Section \ref{sec:res&dis}, we present the results of proposed deep learning models, and we also demonstrate how to use model interpretation to understand hidden patterns learned by the best performing model. Furthermore, in the same section, we explain our results on predictive bio-marker analysis and additional analysis we conducted to evaluate the robustness of the proposed approach.  Finally, in Section \ref{sec:concl}, we summarize our findings.

\section{Related work \label{sec:relatedw}}

The literature on deep learning-based epileptic seizure prediction can be divided into two key branches: patient-independent seizure prediction and patient-dependent seizure prediction. Clearly, developing patient-independent classifiers can be recognized as the complex task since researchers have to handle patient variations and distinguish between seizure-related patterns in the data \cite{Tsiouris2017DiscriminationDatab,Khan2018FocalNetworks}.

The recent study by Tsiouris~et~al. \cite{Tsiouris2017DiscriminationDatab} proposed a Support Vector Machine classifier to identify pre-ictal and interictal brain states from EEG signals. They have employed time domain, frequency domain, and graph theory-based features for training their model. Their final classifier, which was evaluated on all 24 subjects from the CHB-MIT EEG dataset \cite{Shoeb2009ApplicationTECHNOLOGY} achieved an accuracy of 68.50\% for patient-independent classification. The deep learning model proposed by Khan~et~al.~\cite{Khan2018FocalNetworks} is the state-of-the-art model for patient-independent epileptic seizure prediction, and it achieves 0.8660 ROC-AUC score for pre-ictal state detection with a 10 minute early prediction window. This prediction window was determined by the authors as the location where the adopted features start to change when the brain is shifting from  the pre-ictal to interictal state in a majority of subjects. 

Compared to patient-independent studies, a large number of studies can be found investigating patient-specific seizure classification, as a result of the high inter subject variability in EEG data. The recent study conducted by Daoud and  Bayoumi \cite{Daoud2019EfficientLearning} achieves almost 100\% performance for a classifier trained separately for eight patients (from 24) using an Encoder-Decoder CNN + Bidirectional LSTM network. They have used an Encoder-Decoder network as the feature extractor, and have proposed a channel selection algorithm to achieve improved performance.  In another recent investigation, Zhang et al. \cite{artifical} proposed a CNN model  for patient-specific seizure prediction. To overcome the data limitation problem, they have used multi-segment cutting and splicing method and a generative adversarial network for to synthesise new data. In their study they have emphasised that using such an augmentation methods adds complexity and increases the training time of the overall process. Their classifier achieved 92.2\% accuracy with a 30 minutes early prediction window for 23 subjects from the CHB-MIT EEG dataset \cite{Shoeb2009ApplicationTECHNOLOGY}. A similar study that used data augmentation techniques to design patient-specific models can be found in \cite{dataaugment}. 

Another algorithmic method to overcome data scarcity is extracting multiple descriptive features from the EEG waveforms that contain seizure related characteristics. The study by Tsiouris~et~al.~\cite{Tsiouris2018ASignals} presents such an evaluation on patient-specific models and employs the CHB-MIT EEG dataset \cite{Shoeb2009ApplicationTECHNOLOGY}.  Some of the features they considered include time-domain statistical, spectral power-based, autocorrelation-based, and graph theory-based features. They have been able to achieve 99.84\% average accuracy using a Long-Short Term Memory (LSTM) network. 

As noted, to overcome data scarcity issues when training patient-specific deep learning models, researchers have used data-level and feature-level improvement strategies. The problem of high inter subject variability in EEG data has been acknowledged and discussed in the seizure prediction literature, yet few researchers have considered ways of overcoming this problem by designing patient-independent models. Similar concerns have been raised in the recent review on deep learning-based EEG analysis by Roy et al.~\cite{Roy2019DeepReview} who observe that patient-specific models designed for EEG applications demonstrate good performance, since the data has low variability considering a subject, whereas designing patient-independent models is a challenging task. They further note that, examining a broad range of EEG-related studies which have investigated both approaches, patient-specific approaches have often shown better performance than patient independent methods \cite{Roy2019DeepReview}. According to their observations on current EEG-based deep learning applications, there has been a clear trend in investigating patient-independent models recognizing their advantages as real world solutions for EEG applications. 

In summary, we recognize the following limitations in the current epileptic seizure prediction literature.

Little attention has been given to patient-independent studies considering the high inter subject variability in EEG data. As in \cite{Roy2019DeepReview}, we argue that designing patient-independent models is a challenging task, but conducting research related to such models is important as they offer a more realistic solution to the problem.  In the context of this objective, the prediction accuracy and prediction horizon of such methods should be improved considering the predictive capabilities of patient-specific studies. 

Other observed limitations are related to the explainability of the model and the predictive capacity of the input feature representations used. The recent review by Rasheed~et~al~\cite{Rasheed2020MachineReview} discusses the importance of interpretability of the design model. We observe that model interpretation can be an effective tool to build understanding and trust among end-users (who do not necessarily possess a machine learning background) regarding the generated predictions. Furthermore, according to recent review by Kuhlmann et al. \cite{Kuhlmann2018SeizureEra}, deep learning-based seizure prediction studies haven’t focused on the predictive capabilities of the input features used. Therefore, acknowledging these limitations further investigations should be conducted.

\begin{figure*}[t!]
    \centering
    \includegraphics[width=0.8\linewidth]{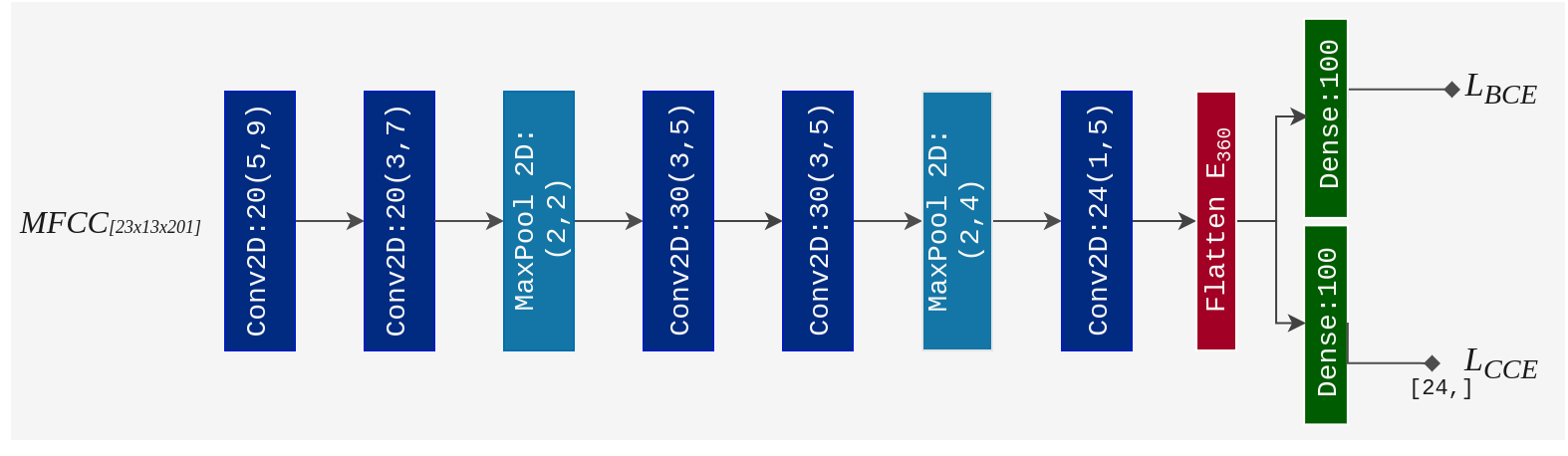}
    \caption{CNN network architecture. $L_{BCE}$: Binary Cross Entropy Loss, $L_{CCE}$ Categorical Cross Entropy Loss. Conv2D:XX(h,w) refers to a 2D Convolution layer with XX (h,w) sized filters. Dense:XX: A fully connected linear layer with XX neurons.}
    \label{fig:cnn-arch}
\end{figure*}

\begin{figure*}[t!]
    \centering
    \includegraphics[width=0.9\linewidth]{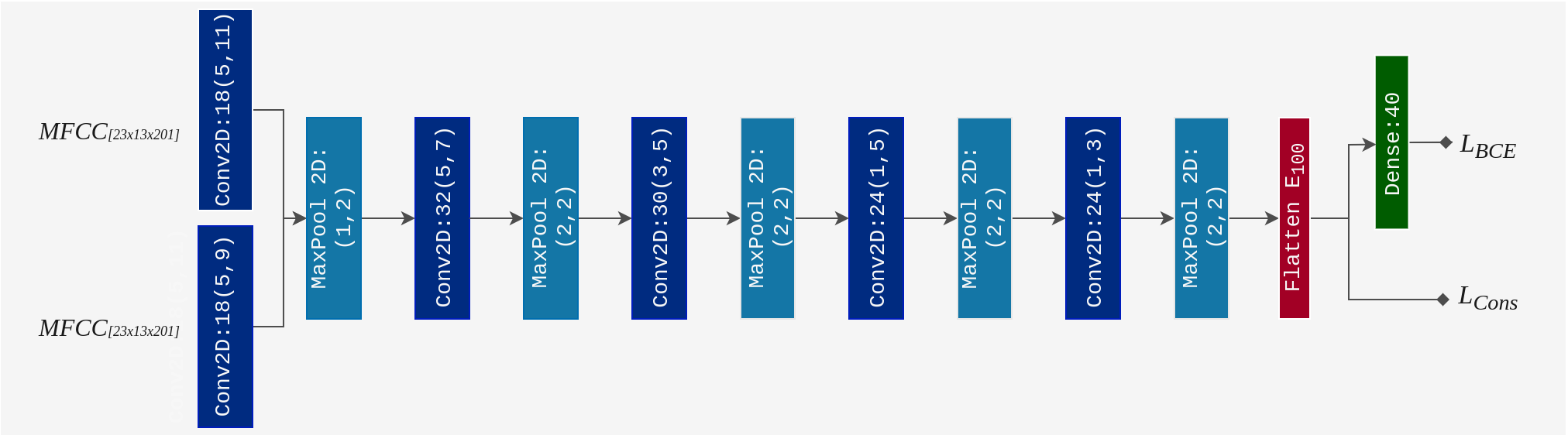}
    \caption{Siamese network architecture. $L_{BCE}$: Binary Cross Entropy Loss, $L_{Cons}$ Categorical Cross Entropy Loss. Here, we only show a one CNN-encoder branch of the Siamese network with the extended classification output.  Conv2D:XX(h,w) refers to a 2D Convolution layer with XX (h,w) sized filters.  Dense:XX: A fully connected linear layer with XX neurons. }
    \label{fig:siamese-arch}
\end{figure*}

\section{Methodology \label{sec: methlgy}}

In this study, we propose two deep learning models for patient-independent epileptic seizure prediction inspired by the novel multitask learning techniques \cite{Caruana1997MultitaskLearning,Zhang2017ALearning}, which allow us to train models that have the ability to recognize differences within patients while also performing seizure classification. Here, we argue that understanding patient differences during training is key to enabling the model to learn complex patterns in the data, which ultimately leads to higher prediction accuracy. The remainder of this section is organized as follows. In Section \ref{susec: data}, we discuss the dataset used for the evaluation. In Sections \ref{subsec: multicnn} and \ref{subsec: siamese}, we propose two different deep learning models for patient-independent seizure prediction. In Section \ref{susec: transfer}, we discuss additional studies we conducted to evaluate the effectiveness of the proposed patient-independent learning technique. 

\subsection{Data Preparation \label{susec: data}}

We use recordings from the CHB-MIT EEG dataset collected at the Children’s Hospital Boston \cite{Shoeb2009ApplicationTECHNOLOGY}. This dataset contains 24 EEG recording samples captured from 23 subjects: 5 male subjects aged between 3–22 years, 17 female subjects aged between 1.5–19 years and an additional non-annotated sample was added later. The additional sample was later captured from subject 1 after 1.5 years, and recent studies have treated this as a separate recording captured as a different subject (i.e yielding 24 cases/subjects) \cite{seizuredetection,Tsiouris2017DiscriminationDatab}. The dataset was intended to be recorded as a continuous EEG database. Therefore, for each subject, it contains annotations related to the capture start times and end times. However, there are some samples with gaps from 10s to multiple hours as a result of hardware limitations.  

Furthermore, some subjects contain 22 EEG-channel recordings from the 10-20 system. To be consistent with other cases, we added an additional channel for those subjects by computing the average across the available 22 channels. We followed this step to ensure the consistency of the dataset, as dropping 1 channel from the 23-channeled EEG spectrum may remove some important seizure related patterns in data.

Similar to the investigation by Daoud and Bayoumi \cite{Daoud2019EfficientLearning}, we assume the pre-ictal state duration of each patient to be at least one hour before the seizure onset, and the interictal state of the brain occurs four hours before or after the seizure onset. Given that the pre-ictal state of the brain is subjective, and it may appear hours before the seizure encounter, this criterion will ensure that we capture the most promising samples to accurately represent the interictal brain state of the subject. After selecting pre-ictal and interictal samples from each patient, we obtained a 158,902 sample balanced dataset where each sample has a duration of 10s and 23-channels. When constructing this balanced dataset,  since the original data has a limited number of pre-ictal signals (or recordings with seizures) we used 2s overlap while windowing those signals. However, our interictal samples are non-overlapping 10s signals and the balanced dataset contain samples from all available EEG recordings in the database. 

For training deep learning models, we used the Mel Frequency Cepstral Coefficients (MFCCs) of the sampled signal. This selection is based on MFCC's wide range of applicability for biomedical signal-related deep learning tasks \cite{Dissanayake2020UnderstandingDetection}. To compute the spectrum, we used 13 Filter Banks within the frequency range of 0-256.0Hz, resulting in a feature map of shape $[\textbf{23}\times13\times201]$ (for \textbf{23} channels). It should be noted that, similar to \cite{Khan2018FocalNetworks}, we use a 10-Fold Cross-Validation to evaluate performance, in each fold our deep learning model will be trained on $\approx$140k instances and be validated on $\approx$15k instances.

\subsection{Multitask Deep Learning Architecture \label{subsec: multicnn}}

In this section, we introduce a multitask Convolution Neural Network for recognizing pre-ictal brain states. The proposed CNN is able to differentiate between patients and pre-ictal-interictal brain classes. As shown in Figure \ref{fig:cnn-arch}, the proposed model has two outputs: a seizure-related binary classification output and a patient prediction output of size $[24\times1]$. Considering the architecture of the model, it has five 2D convolution layers with different filter sizes. Furthermore, for regularization, we use Dropout with 0.6 probability and 0.4 MaxNorm kernel normalization constraint for each convolution layer. 
  
\begin{equation}
\label{eq:loss-1}
    Loss = \lambda~L_{BCE} + (1 - \lambda)~L_{CCE}.
\end{equation}

As shown in Equation \ref{eq:loss-1}, we employ a combined loss function to train the model. Here, $L_{BCE}$ refers to the Binary Cross-Entropy loss for the seizure classification task, $L_{CCE}$ refers to the Categorical Cross-Entropy loss for the patient prediction task, and $\lambda$ is a hyperparameter that determines the contribution from individual losses and is determined experimentally.

\subsection{Siamese Architecture \label{subsec: siamese}}

The model discussed in Section \ref{subsec: multicnn} implements a deep learning architecture that is able to learn two different but related concepts in the data. Examining recent literature \cite{Motiian2017UnifiedGeneralization}, Siamese networks also demonstrate an ability to learn such complex patterns. Therefore, in this section, we evaluate how efficiently Siamese networks can be utilized as pre-ictal/interictal brain state classifiers.

Figure \ref{fig:siamese-arch} illustrates one classification branch of the proposed Siamese network architecture for computing a unique embedding for each patient (of shape $[100\times1]$). The model has two convolution channels as the input, which extract features using two different kernels shaped $[5\times9]$ and $[5\times11]$. This model also uses Dropout with 0.4 probability as a regularizer. Since the model also acts a as a pre-ictal-interictal classifier, we use a linear (40 neurons) layer to produce the final classification output.  

\begin{equation}
\label{eq:loss2}
Loss = \gamma~L_{Cons} + (1 - \gamma)~L_{BCE}.
\end{equation}

Equation \ref{eq:loss2} expresses the loss function used for training the Siamese network. Here, $L_{Cons}$ refers to the Contrastive loss \cite{Hadsell2006DimensionalityMapping}, $L_{BCE}$ refers to the Binary Cross Entropy loss, and $\gamma$ is a hyperparameter that controls the relative weight of the terms. The Contrastive loss can be expressed in the following equation where $Y_{true}$ indicates whether the two samples are from the same class, $d$ is the distance measure used and $margin$ is the minimum separation between embeddings of different classes.

\begin{equation}
\label{eq:constrastive}
L_{Cons} = Y_{true}~d^2 + (1 - Y_{true})~max(margin - d, 0).
\end{equation}

We set the margin to 1.0 and we use L2 distance as the distance measure ($d$). Within the Siamese framework, we consider pairs of samples from the same patient to be a matched pair, and pairs of samples from different patients to be mis-matched pairs. For this, offline mining \cite{Hadsell2006DimensionalityMapping} is used to generate paired data to train the network. Binary Cross Entropy loss for seizure classification is computed for the first sample of the pair (see Figure \ref{fig:siamese-arch}). 

For the offline mining technique, the data stream yields pairs of samples for the loss calculation (primary and secondary). To generate this data stream, we use the original dataset. For each unique primary sample in the dataset, we select a second sample such that 50\% of pairs contain samples from the same patient, and 50\% contain samples from different patients.

\subsection{Additional Evaluation: Patient-specific Seizure Prediction \label{susec: transfer}}

\begin{figure*}[t!]
    \centering
    \begin{subfigure}[b]{0.24\textwidth}
        \includegraphics[width=\textwidth]{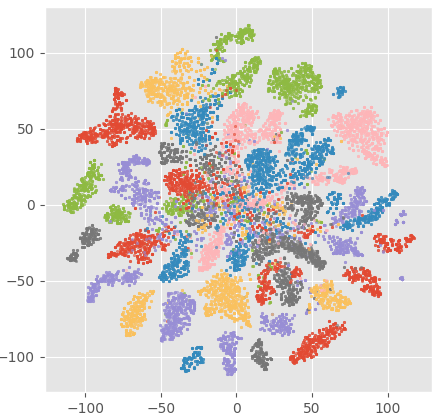}
        \caption{Subjects}
        \label{subf:pat-cnn}
    \end{subfigure}
    \begin{subfigure}[b]{0.24\textwidth}
        \includegraphics[width=\textwidth]{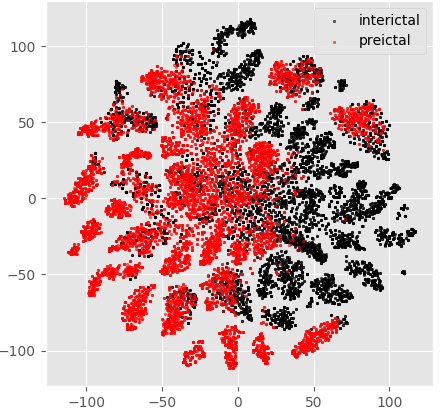}
        \caption{Pre-ictal interictal samples}
        \label{subf:clsf-cnn}
    \end{subfigure}
    \begin{subfigure}[b]{0.24\textwidth}
        \includegraphics[width=\textwidth]{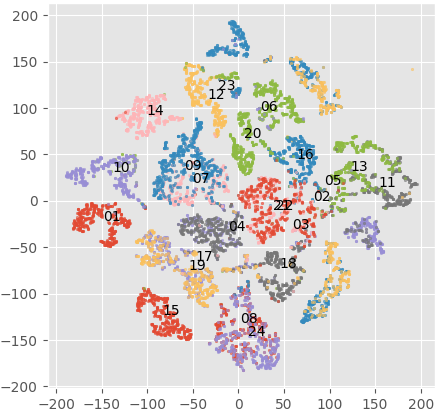}
        \caption{Subjects}
        \label{subf:pat-siamese}
    \end{subfigure}
    \begin{subfigure}[b]{0.24\textwidth}
        \includegraphics[width=\textwidth]{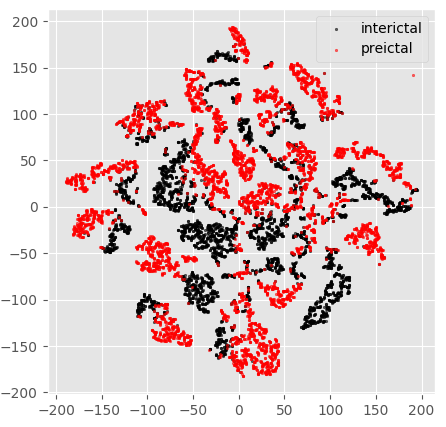}
        \caption{Pre-ictal interictal samples}
        \label{subf:clsf-siamese}
    \end{subfigure}
    \caption{t-SNE \cite{vanDerMaaten2008} visualizations for the embeddings from models introduced in Sections \ref{subsec: multicnn} (Figs. a, b) and \ref{subsec: siamese} (Figs. c, d). Each visualization shows 400 randomly selected instances from each subject.}
    \label{fig: tsne}
\end{figure*}

As mentioned above, the primary focus of this study is developing a patient-independent seizure prediction model. We also conduct an additional evaluation regarding designing subject-specific classifiers using Transfer Learning. 

As discussed in Section \ref{sec:relatedw}, the number of data instances available for a particular patient is a major limitation when designing patient-specific seizure classifiers. Therefore, to address this problem, we use Transfer Learning to transfer information learnt in patient-independent models  to patient-specific models.
It should be noted that our dataset does contain a considerable amount of samples from each patient (all 24), but those samples are not sufficient to train a deep learning classifier from scratch. Therefore, in this experiment, we demonstrate how we can use Transfer Learning to achieve state-of-the-art patient specific classification from the learned model from Section \ref{subsec: siamese} (i.e the Siamese-based classifier). This strategy can be seen as the most data efficient way to solve the problem compared to previous studies \cite{Daoud2019EfficientLearning,Tsiouris2018ASignals}.

In this experiment, first we remove instances ($d_i$) extracted from the selected subject $i$ ($i \in [1,2,\dots,24]$) from the dataset (both training and evaluation). Then, we train the model ($M_i^{'}$) using the rest of the available data. 
After training the model, we transfer the learned model ($M_i^{'}$) to the space of subject $i$ by re-training (or fine-tuning) the model utilizing the instances ($d_i$) from subject $s_i$. We perform this evaluation for all 24 subjects, and train models for  200 epochs.
As an additional task, we evaluate the performance of Transfer Learning by varying the number of selected training examples from $d_i$. Here, to compare the classification accuracy, we use the same validation dataset selected from $d_i$. By doing this, we analyse the performance variation of the model as the number of instances available changes.

\section{Results and Discussion \label{sec:res&dis}}

The results and discussion is organized as follows. In Section \ref{res:sec1}, we discuss the accuracies obtained by proposed learning techniques introduced in Sections \ref{subsec: multicnn} and \ref{subsec: siamese}. In this section, we also use t-SNE \cite{vanDerMaaten2008} analysis to visualize how those two methods represent the patient differences in the data while acting as pre-ictal-interictal state classifiers. In the next section we demonstrate how we can use model interpretation techniques to understand what sort of input features in the MFCC map contribute to the final prediction. Here, we only focus on channel-level attribution of the input feature maps of the Siamese network. Next, we employ a KL-Divergence-based probabilistic technique to determine the exact point where the brain state changes from the interictal state to the pre-ictal state using MFCC features. The final section presents results from the additional evaluation discussed in Section \ref{susec: transfer}

\begin{figure*}[t!]
    \centering
    \includegraphics[width=0.65\linewidth]{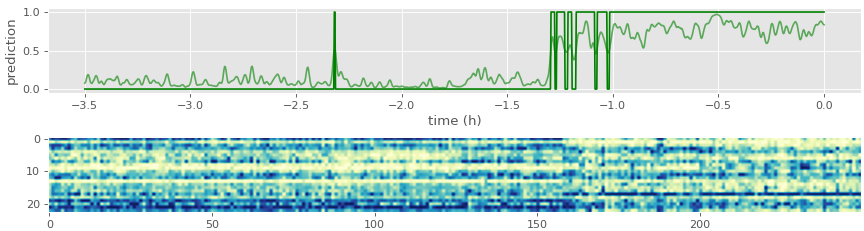}
    \caption{Prediction (top) and Channel-Shapley value variation (bottom) for the seizure sample chb06\_09.edf. Shapley values with higher contributions are indicated in a darker blue color.
    Top figure shows the actual prediction and the final prediction (thresholded).}
    \label{fig:chb06.edf}
\end{figure*}

\begin{figure*}[t!]
    \centering
    \includegraphics[width=0.80\linewidth]{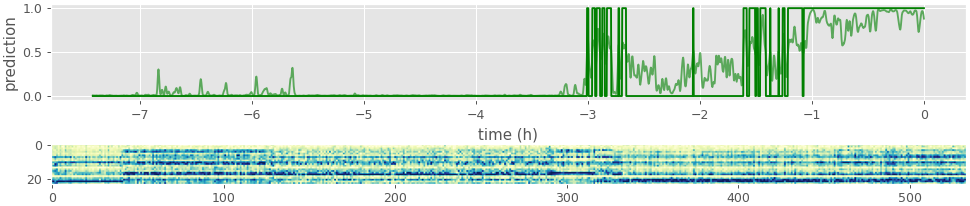}
    \caption{Prediction (top) and Channel-Shapley value variation (bottom) for seizure samples chb09\_05.edf and chb09\_06.edf. Shapley values with higher contributions are indicated in a darker blue color. Top figure shows the actual prediction and the final prediction (thresholded).}
    \label{fig:chb9.edf}
\end{figure*}

\begin{figure*}[t!]
    \centering
    \includegraphics[width=0.72\linewidth]{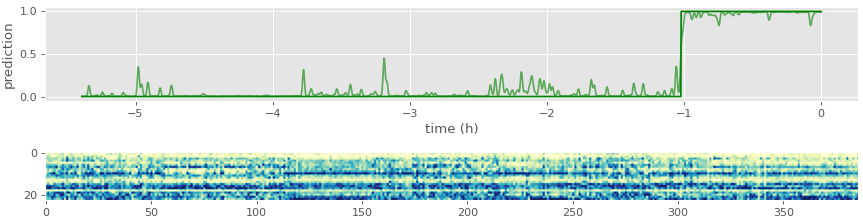}
    \caption{Prediction (top) and Channel-Shapley value variation (bottom) for seizure samples chb07\_11.edf and chb07\_12.edf. Shapley values with higher contributions are indicated in a darker blue color. Top figure shows the actual prediction and the final prediction (thresholded).}
    \label{fig:chb07.edf}
\end{figure*}

\begin{figure*}[t!]
    \centering
    \includegraphics[width=0.65\linewidth]{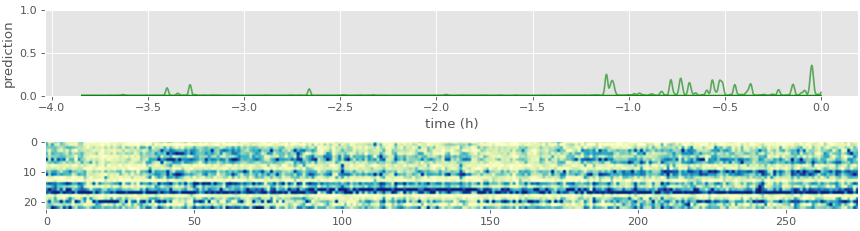}
    \caption{Prediction (top) and Channel-Shapley value variation (bottom) for the normal sample chb04\_17.edf. Shapley values with higher contributions are indicated in a darker blue color. Top figure shows the actual prediction and the final prediction (thresholded).}
    \label{fig:chb04.edf}
\end{figure*}

\begin{table}[b!]
    \centering
    \begin{tabular}{|c|c|c|c|}
    \hline
    Duration (minutes)  & 15 & 30 & 60 \\ \hline
    Accuracy & 95.72\%  & 94.72\% & 91.54\%  \\ 
    ROC-AUC & 0.9877 & 0.9843 & 0.9694 \\
    Sensitivity & 97.88\%  & 96.43\% & 92.45\% \\ \hline
    \end{tabular}
    \caption{Results after changing the pre-ictal sample duration selected for training the Siamese Network. Overlaps in pre-ictal samples:15mins : 3.5s,  30mins : 2.5s}
    \label{tab:seitime}
\end{table}

\subsection{Patient Independent Seizure Prediction\label{res:sec1}}

As previously mentioned, we set the prediction horizon as one hour, and our dataset contains 10s 23-Channel EEG samples from all 24 subjects. We trained our models for 200 epochs with a batch size of 600. We used the Adam optimizer with a 0.001 learning rate. 

The following list demonstrates the 10 Fold Cross Validation results for our two models (Model 1 from Section \ref{subsec: multicnn} and Model 2 from Section \ref{subsec: siamese}), 

\begin{itemize}
    \item \textbf{Model I:} Accuracy: pre-ictal-interictal state detection 88.81($\pm$0.27)\%, patient detection (additional task): 95.00($\pm$0.56)\%. Sensitivity: 93.45($\pm0.22$)\%, Specificity: 81.64($\pm0.23$)\%, ROC-AUC score: 0.9273($\pm$0.0029), $\lambda$ =  0.9
    
    \item \textbf{Model II:} Accuracy: pre-ictal-interictal state detection 91.54($\pm$0.17)\%, Sensitivity: 92.45($\pm0.22$)\%, Specificity: 89.94($\pm0.21$)\%. Constrastive loss: 0.0475($\pm$0.002), ROC-AUC score: 0.9694($\pm$0.0018), Optimal Embedding size: 100, $\gamma$ = 0.6 
\end{itemize}

As discussed in previous sections, both models in the investigation have been trained as multitask models to predict seizure and patient-related information. In this setting, the Siamese network from Section \ref{subsec: siamese} shows superior performance. However, considering that the models incorporate data from a one-hour prediction window, both models perform better than the state-of-the-art model discussed in \cite{Khan2018FocalNetworks}, which achieves 0.8660 ROC-AUC score for a 10 minute prediction window. We also note \cite{Khan2018FocalNetworks} only considered  15 subjects from the same dataset, while models proposed in our investigation shows higher accuracies on a more diverse dataset with 24 subjects. The classifiers also achieve significant accuracy gains compared to studies by \cite{Tsiouris2017DiscriminationDatab} and \cite{Sridevi2019ImprovedSeizuresf}.

Examining the proposed models, both architectures consist of a CNN encoder network followed by an additional linear layer(s). Looking at the number of trainable parameters in each classifier: the CNN multitask model contains 102k parameters and the Siamese network holds 128k parameters. Even though the proposed models have similar architectural arrangements, they employ two different strategies to learn, and the intermediate embeddings produced by both models differ (it should be noted that the lengths of those two embeddings were experimentally chosen to achieve high classification accuracies). We use t-SNE \cite{vanDerMaaten2008} to understand the capability of those embeddings. By doing this, we seek to understand how effectively each of these embeddings interprets patient and seizure-relevant information. 

Figure \ref{fig: tsne} shows four t-SNE \cite{vanDerMaaten2008} visualizations, illustrating how each of the intermediate embeddings represents the concepts in the data in a 2D plane. Figures \ref{subf:pat-cnn} and \ref{subf:clsf-cnn} present t-SNE \cite{vanDerMaaten2008} visualization of the ($[360\times1]$) sized embedding from the CNN model. The next two figures use the $[100\times1]$ embedding from the Siamese network. Here, for each selected embedding, first we show the subject distribution (different colours represent different subjects) and then we demonstrate how pre-ictal and interictal samples are distributed in the same space.

According to Figures \ref{subf:pat-cnn} and \ref{subf:pat-siamese}, compared to the CNN-based embedding, the embedding generated from the Siamese network shows comparatively good results in separating patient-related information in the data.  In most cases, for a particular patient,  the Siamese embedding yields a smaller number of sub-clusters than the CNN-based embedding (see specific color codes). Furthermore, observing class separations in Figures \ref{subf:clsf-cnn} and \ref{subf:clsf-siamese}, the CNN-based embedding appears to clearly divide class information in the higher dimensional space by producing multiple clusters. In contrast, rather than appearing as sub-clusters in the embedding space, class-related patterns in Siamese embeddings seem to be distributed within the patient-clusters. In fact, that behavior itself forces the Siamese model to have a smaller number of clusters. 

Analysing Figure \ref{fig: tsne}, one of the reasons for the Siamese model showing superior performance is it's ability to clearly separate patients. Hence, knowing the patient differences while being trained as a pre-ictal-interictal state classifier seems to be the key to achieving higher seizure prediction performance. Recent patient-independent seizure prediction architectures in the literature have overlooked this, and as mentioned,  this may be one of the reasons for having considerably low performance compared to patient-dependent seizure models in \cite{Daoud2019EfficientLearning,Tsiouris2018ASignals}. 

\begin{figure*}[t!]
    \centering
    \includegraphics[width=1.0\linewidth]{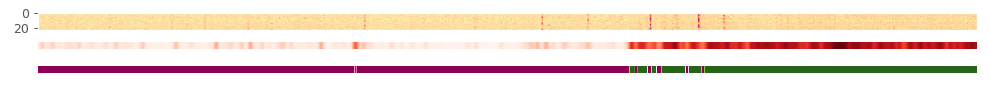}
    \includegraphics[width=0.6\linewidth]{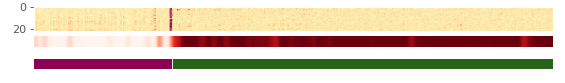}
    \includegraphics[width=1.0\linewidth]{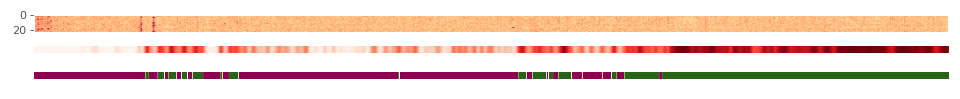}
    \includegraphics[width=0.85\linewidth]{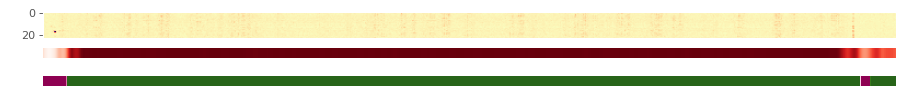}
    \includegraphics[width=0.52\linewidth]{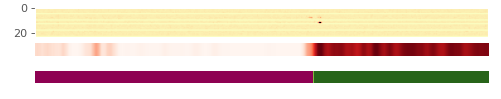}
    \caption{Visualization for the KL-Divergence variation of Channel-Level input feature maps ($[13\times201]$) computed for a sample with a seizure onset (chb06\_09.edf,~chb07\_12.edf,~chb09\_06.edf,~chb04\_05.edf and chb22\_11.edf). In each plot, darker purple color spots indicates a higher distribution shift. We also indicate the prediction of model $(0.0,1.0)$ in a red colorbar, and the final (thresholded) prediction \textit{purple} interictal, \textit{green} pre-ictal.}
    \label{fig:kldiv}
\end{figure*}

Table \ref{tab:seitime} shows cross validation results when varying the pre-ictal duration taken to create the dataset. As shown in the table, the proposed Siamese network demonstrate higher performance when evaluated on shorter pre-ictal durations (i.e closer to the seizure onset).

\subsection{Model Interpretation}

The SHAP (SHapley Additive exPlanations) by Lundberg and Lee \cite{shaplung} is one of the most popular model interpretation techniques in the deep learning literature. The SHAP algorithm provides insights related to the input feature contributions for a particular prediction (i.e attribution). Simply, the SHAP algorithm can be expressed as the following linear regression equation where $x_i$ refers to the $i^{th}$ feature, $w_i$ refers to the weight associated with the $i^{th}$ feature and $\hat{y}$ is the prediction made by the model.  

\begin{equation}
\label{shap}
 \hat{y} = w_{0}x_{0} + w_{1}x_{1} + w_{2}x_{2} + \dots + w_{n}x_{n}. 
\end{equation}

In this context, a Shapley value determined by the SHAP algorithm denoted by $w_i$, is an indication of the contribution of feature $x_i$ to the final prediction. Furthermore, Shapley values can be negative or positive symbolizing the direction of influence.  For simplicity, in this investigation, we only use absolute Shapley values such that a larger value indicates a greater contribution. 

Since the input data to the network is high dimensional, we use a channel-level interpretation strategy to discuss how each channel contribute to the final prediction made by the model. To demonstrate this, we use six samples from four different patients from the CHB-MIT EEG dataset. In our visualizations, Figures \ref{fig:chb06.edf}, \ref{fig:chb9.edf}, and \ref{fig:chb07.edf} are samples with seizures, and Figure \ref{fig:chb04.edf} is a normal sample selected from subject four. Furthermore, samples shown Figures \ref{fig:chb06.edf} and \ref{fig:chb07.edf} are combined samples with less than 60s between their capture (see specific file names in the figure). In this analysis, we only consider the Siamese network proposed in Section \ref{subsec: siamese}, which demonstrates superior performance compared to the CNN model. 

For a given instance $k_{[23\times13\times201]}$ in the input dataset, the SHAP algorithm returns a Shapley value map with the same shape. To compute channel-level attribution ($C_{[23\times1]}$), we use Equation \ref{eq:ch-shap},

\begin{equation}
\label{eq:ch-shap}    
    C_{[23\times1]} = \sum_{i = 0}^{i = 13\times201} |k_{ji}|,~~\forall~j \in [1,2,\dots,23].
\end{equation}

Each plot in the bottom of Figures \ref{fig:chb06.edf} to \ref{fig:chb04.edf} illustrates the channel-level Shapley value map shaped $23\times N$ (N: the number non overlapping of 10s MFCC input feature instances). Here, we indicate higher Shapley values (i.e. greater contribution) in darker shades of green. The top figure of each plot is the prediction made by the model for those instances before the seizure onset (the selection period varies from 3.5 hours to 7.0 hours). For clear visualization of  the interpretation, we apply a Hanning smoothing window \cite{scipy} to the predictions.  

Figure \ref{fig:chb06.edf} demonstrates the channel-level Shapley value variation with the prediction made by the model (exact and thresholded) for subject six. According to the prediction, the pre-ictal brain state of subject six appears to be visible 1.25 hours (1 hour 15 minutes) before the seizure onset. Examining the SHAP variations, the highest contributing EEG channel appear to change at point at which the brain state switches from interictal to pre-ictal brain state. Furthermore, looking at the overall Shapley values, almost all channels appear to show some contribution to the final prediction. 

Figure \ref{fig:chb9.edf} shows the interpretation visualization for the combined EEG signal from subject nine. As per the previous interpretation, this also shows similar channel transfers at point at which the brain state shifts. However, the pre-ictal stage seems to be appear three hours before the seizure.  

Compared to previous results, the prediction for patient seven shown in Figure \ref{fig:chb07.edf} has a steady shift between interictal and pre-ictal brain states. Furthermore, this figure does not seem to show the channel shifting behavior observed in the previous two explanations.

The final sample taken for the interpretation is from patient four in the CHB-MIT EEG dataset. This interpretation visualization shows how the model classifies a seizure-free sample. Examining Shapley values, a single channel (channel 18) seems to contribute prominently to the prediction made by the model. Unlike previous interpretations, this clearly shows an individual channel from the EEG signal as the primary contributor for the prediction.

Collectively examining the results, the Siamese model proposed in this study has a steady pre-ictal-interictal brain state recognition capability. Looking at the predictions made for patients six and nine, the pre-ictal brain state appears to be visible 1.5 hours before the seizure onset. 
Moreover, observing Shapley values for each channel, it is apparent that almost all channels in the input feature map do contribute to the prediction made by the model.  

\subsection{Pre-ictal bio-markers derived from MFCC feature maps. \label{subsec: biomek}}

\begin{table*}[t!]
    \centering
    \begin{tabular}{|c|c|c|c|c|c|c||c|c|c|c|c|c|c|}
    \hline
    P & N & LOPO & $100$ &$1000$ & $2000$  &$N$ & P & N & LOPO &$100$ & $1000$ & $2000$ & $N$ \\ \hline

    chb01 & 6905 & 60.01 & 92.14 & 96.62 & 97.91 & 99.50 & 
    chb13 & 5989 & 63.11 & 95.17 & 97.49 & 98.27 & 99.62 \\ 
    
    chb02 & 5283 & 58.36 & 88.82 & 94.63 &  97.85 & 98.63 & 
    chb14 & 6523 & 61.58 & 75.57 & 79.38 & 81.48  & 91.17 \\
    
    chb03 & 6255 & 49.75 & 85.88 & 97.10 & 97.10  & 99.20 & 
    chb15 & 7734 & 37.21 & 69.88 & 81.22 & 83.57 & 97.69 \\
    
    chb04 & 16876 & 78.72 & 85.50 & 92.58 & 93.91 & 98.75 & 
    chb16 & 3661 & 53.99 & 83.10 & 89.92 & 92.21 & 96.71 \\
    
    chb05 & 6089 & 55.36 & 62.89 & 80.81 & 85.32 & 94.63 & 
    chb17 & 4515 & 41.55 & 85.10 & 96.43 & 97.28 & 97.28 \\
    
    chb06 & 9542 & 59.56 & 61.81 & 74.45 & 75.50 & 93.11 & 
    chb18 & 6390 & 56.04 & 80.44 & 97.12 & 96.90 & 96.98 \\
    
    chb07 & 9042 & 61.22 & 81.10 & 90.78 & 92.75 & 98.92 & 
    chb19 & 4697 & 56.71 & 98.30 & 100.0 & 100.0 & 100.00 \\
    
    chb08 & 5749 & 68.38 & 86.46 & 92.52 & 91.41 & 96.65 & 
    chb20 & 4580 & 48.08 & 97.34 & 98.17 & 99.56  & 99.56 \\
    
    chb09 & 8958 & 62.79 & 82.27 & 90.20 & 92.33  & 97.59 & 
    chb21 & 5567 & 64.05 & 73.21 & 86.65 & 87.56 & 88.56 \\
    
    chb10 & 10480 & 58.50 & 81.43 & 87.92 & 92.96 & 98.01 & 
    chb22 & 5251 & 60.52 & 81.45 & 84.26 & 87.04  & 87.04 \\
    
    chb11 & 4513 & 51.64 & 90.89 & 97.00 & 97.65 & 99.00 & 
    chb23 & 3847 & 73.59 & 93.33 & 98.17 & 98.10 & 98.10 \\
    
    chb12 & 5804 & 69.81 & 88.15 & 92.50 & 92.33 & 96.68 & 
    chb24 & 4662 & 53.77 & 80.00 & 88.33 & 88.55  & 88.55 \\ \hline \hline
    
    \multicolumn{9}{|c|}{Average Accuracies:} &
     58.55 & 83.34 & 91.01 & 92.39 & 96.67 \\ \hline
    
    \end{tabular}
    \caption{Validation results from the Transfer Learning approach for patient-dependent seizure prediction (N: the number of samples). Batch sizes for the selected samples (sample\_size\textbf{:}batch\_size): $100:10$, $1000:100$, $2000:200$, $N:400$. Leave One Patient Out (LOPO) column presents the validation accuracies before transferring the model.} 
        \label{tab:transfer}
\end{table*}

Recognizing one of the limitations discussed in \cite{Kuhlmann2018SeizureErab}, in this section, we investigate what sort of predictive characteristic (or feature) appears when the brain state shifts from the pre-ictal to interictal stage. This analysis will help to localize the exact point where the feature distribution changes, and will help to determine if such a detail is visible in the MFCC feature space rather than in the hidden space of the deep learning model. 

As in the previous section, we use channel-level evaluations due to the high dimensionality of the input data. First, we represent a particular input feature channel $F_{t}$ at a given time $t$ as a probability distribution $P_t$, and in this setting, $P_{t}(f^k_t)$ is the probability of having feature value $f^k_t$ in the distribution ($f^k_t \in F_t$). Next, we define the change of the feature map due to the brain-state-shift as the KL-Divergence between $P_{t}$ and $P_{t+1}$. Then, we compute this for all time steps in a given channel, and the resulting computation for all 23 channels has the shape $23\times~N$ (N: number of time steps). 

Figure \ref{fig:kldiv} illustrates the results after calculating these variations for five seizure samples. Here, each visualization contains three rows. The top row presents the KL-Divergence-based map computed. The middle row represents the prediction made by the model within the range $(0,1)$, where a darker red color indicates prediction of a pre-ictal brain state. The third row shows the final prediction of the model, which is taken by thresholding the prediction shown in the previous map. Here, green indicates pre-ictal brain states.

Examining all visualizations, it is apparent that the input feature maps shows a clear change when shifting brain state from the interictal state to the pre-ictal state. Furthermore, in the top three figures, all channels in the EEG seem to have changing characteristics, whereas the rest of the explanations only show variations in a single-channel. 

Observing the MFCC feature distribution variation, the designed deep learning classifier does identify a shifting point, and the identified point may be the actual pre-ictal brain state start time/location. Even though this result aligns with the prediction made by the model, we believe that the physiological characteristics related to this observation should be further investigated on a larger database. Importantly, this evaluation implies that such a feature exists in the data, and it can be determined through a simple probabilistic evaluation in the input space. 

Furthermore, given that we can observe such variations for all five subjects, it is apparent that the MFCC feature does offer a unified higher dimensional representation of the EEG signals that helps to differentiate between pre-ictal and interictal shifts. In fact, that point itself demonstrates the validity of the patient-independent model, and the strength of MFCC features used.

\subsection{Additional Evaluations: Seizure prediction: Patient-Specific.}

Examining recent studies on patient-specific seizure prediction, we observed that some of those studies restricted their evaluations to fewer patients due to the lack of seizure recordings. Therefore, in this section, we report the results from Section \ref{susec: transfer}, where we introduced Transfer Learning as a method for addressing data limitations. It should be noted that this is an additional evaluation we conducted along with our main objective i.e designing a patient-independent model for epileptic seizure prediction. 

Table \ref{tab:transfer} presents the results after applying transfer learning to all 24 patients in the dataset. As discussed, we show the Transfer Learning accuracy variation as the number of samples used for transferring the model changes. Table \ref{tab:transfer} also presents the validation accuracy obtained from the entire dataset before transferring the model (i.e a Leave One Patient Out (LOPO) validation). Similar to the previous analysis, we use the Adam optimizer for fine-tuning the model, and we adjust the batch size depending on the number of available training samples (see Table \ref{tab:transfer}). For performance comparison purposes, we keep the same validation dataset for all Transfer Learning evaluations.
Observing the accuracies gained after transferral, the majority of transferred models employing the entire training dataset perform with an accuracy of ~96.67$\pm$3.62\% for patient-specific seizure prediction. Also, the proposed method shows comparatively good accuracy even when transferred using a smaller data sample. Therefore, we believe this demonstrates the realistic and efficient nature of Transfer Learning when designing seizure classifiers for patients with fewer EEG recordings. 

Along with the Transfer Learning evaluation, we test the model's generalization ability by supplying a set of instances from a completely unseen subject (i.e LOPO evaluation).  Here, it should be noted that the problem of patient independence does not deal with generalization aspects \cite{Roy2019DeepReview}, and therefore, generalizing across multiple subjects should be separately investigated. Studying the literature, we were unable to find a deep learning-based patient-independent epileptic seizure prediction study that directly focused on the generalization aspects of the designed model regarding completely unseen patients. However, there are some studies who have used the Leave One Sample Out evaluation method (deals with generalizing to an unseen sample from a subject) \cite{Daoud2019EfficientLearning}, and determining a generalized set of features for training deep learning models on different databases \cite{TRUONG2018104}. However, as in \cite{Roy2019DeepReview,AHMEDTARISTIZABAL201965}, we argue that the generalization capability of the model should be evaluated using completely unseen subjects, and therefore acknowledging this further investigations should be conducted. 

Examining the results, it is apparent that in most cases the model struggles (average: 58.55\%). However, for 11 subjects in the dataset (from 24), the model shows classification capability higher than 60\%, which is a promising result considering the prediction horizon used to design the model. Furthermore, looking at the transfer learning results, even with few data instances, the model is capable of achieving comparable results for solving the problem. 

Recent machine learning models for patient-dependent seizure prediction have shown promising results, and our proposed model also achieves similar accuracies when evaluated on all 24 patients. Considering the number of patients adopted for evaluation and the performance, our model outperforms the recent studies by \cite{Daoud2019EfficientLearning,Zhang2020EpilepsyNetwork,Zabihi2020Patient-SpecificNullclines}.

\section{Conclusion \label{sec:concl}}

The main objective of this study is to design a deep learning classifier for patient-independent epileptic seizure prediction, recognizing its importance for the seizure prediction problems. 

In this research, we proposed two different CNN architectures for this problem. Both proposed models have the ability to accurately recognize seizures with a one-hour prediction window, and those models outperform the state-of-the-art with 7\% and 11\% ROC-AUC score gains respectively. The Siamese model introduced in this study demonstrated excellent performance and we investigated its capability through t-SNE \cite{vanDerMaaten2008} and SHAP model interpretations.

Along with these contributions, we also explained how we can apply the SHAP algorithm as a technique for understanding individual EEG channel contributions. Since this phenomenon has never been studied in the epileptic seizure prediction literature, we believe our study will provide researchers with insight into the importance of model interpretation as a way of understanding the behavior of the model.

As an additional step to evaluate the robustness of our classifier, we showed how to design patient-specific seizure prediction models employing Transfer Learning. The models proposed in our study had 96\% an average  accuracy considering all 24 subjects in the dataset. 

We also conducted an additional analysis regarding the generalization aspect of the model. Given that learning a subject-independent function is itself a complex task, our model showed promising generalization for some completely unseen subjects. As a future direction, we encourage researchers to conduct research related to designing generalized patient-independent classifiers, acknowledging their importance as real-world solutions for the seizure prediction.

\bibliography{Bibs/references,Bibs/bibs}
\bibliographystyle{IEEEtran}

\end{document}